\documentclass{article}
\usepackage{amsmath,graphicx,02460}
\usepackage{float}
\usepackage{booktabs}
\usepackage{bm} 
\usepackage[dvipsnames]{xcolor}
\usepackage{subcaption}
\usepackage{diagbox}
\usepackage{balance}

\title{Predicting transmembrane protein topology from 3D structure}
\name{Author(s) Name(s)\thanks{Thanks to XYZ agency for funding.}}
\address{Author Affiliation(s)}
\twoauthors
 {Sitong Chen}
	{Technical University of Denmark\\
	s230027@dtu.dk}
 {Xiaopeng Mao}
	{Technical University of Denmark\\
	s194408@dtu.dk}

\begin{document}
%

\maketitle
\begin{abstract}
This paper presents a novel approach to infer protein topology using the state-of-the-art graph neural network (GNN), SchNet. The model is trained on the same dataset used to develop the recent DeepTMHMM model with 5-fold cross-validation. Unlike the conventional approaches based on using only the protein sequences or the $\alpha$-carbons as features, we have decoded our classifier in this way, so all atom-level embeddings are used. Without applying any pre-trained weight, the final results have shown great potential that GNNs can be used for topological predictions.

\end{abstract}

\section{Introduction}
\label{sec:intro}

\begin{figure}[H]
    \centering
    \includegraphics[width=\linewidth]{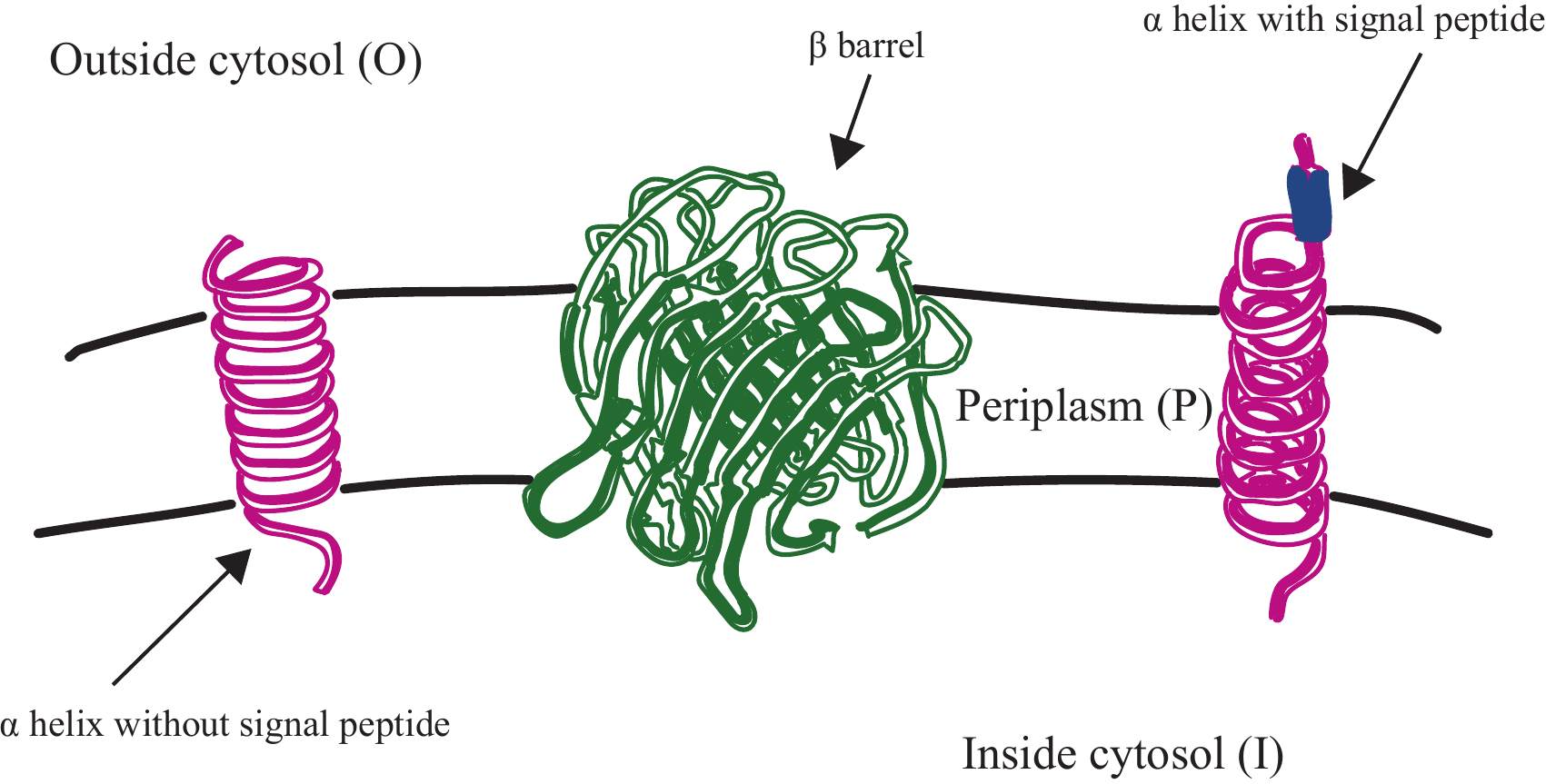}
    \caption{Simple illustration explaining protein topologies. A protein can have different topologies depending on where it is. Please notice that the $\beta$ barrel protein shown in the middle has more complicated topologies as its bottom part frequently changes location. On the other hand, $\alpha$ helices have relatively simpler topologies but they might have signal peptide bound to them, which needs to be distinguished. Illustration inspired by \cite{alberts2015essential}.}
    \label{fig:protein}
\end{figure}

Protein topology tells about the location of a particular protein, which can be inside, outside, or in-between the cell membrane (transmembrane protein) as illustrated in Figure \ref{fig:protein}. A transmembrane protein might serve as a receptor for cell-to-cell communication while a free-moving protein outside or inside a cell might play an important role in cellular transportation \cite{alberts2015essential}. Currently, many machine learning-based models have been developed to predict this very topology by using 1D protein sequences, 3D protein structures, or both features. In \cite{bernsel2009topcons}, Bernsel \textit{et al.} used an ensemble of 5 different 1D sequence machine learning models to predict the protein topology. In another study, Dobson \textit{et al.} achieved a better result using a similar approach but with 10 models instead \cite{dobson2015cctop}. Recently, Hallgren \textit{et al.} and M. Bernhofer \textit{et al.} achieved excellent performances with the DeepTMHMM and TMbed models based on pre-trained protein Language Models (pLMs), respectively \cite{hallgren2022deeptmhmm, bernhofer2022tmbed}. In this study, we present a Graph Neural Network (GNN)-based method that directly uses 3D structural information obtained from the well-known Alphafold model together with an indirect application of the 1D protein sequence.

\section{Methodology}
\begin{figure}[H]
    \centering
    \includegraphics[width=\linewidth]{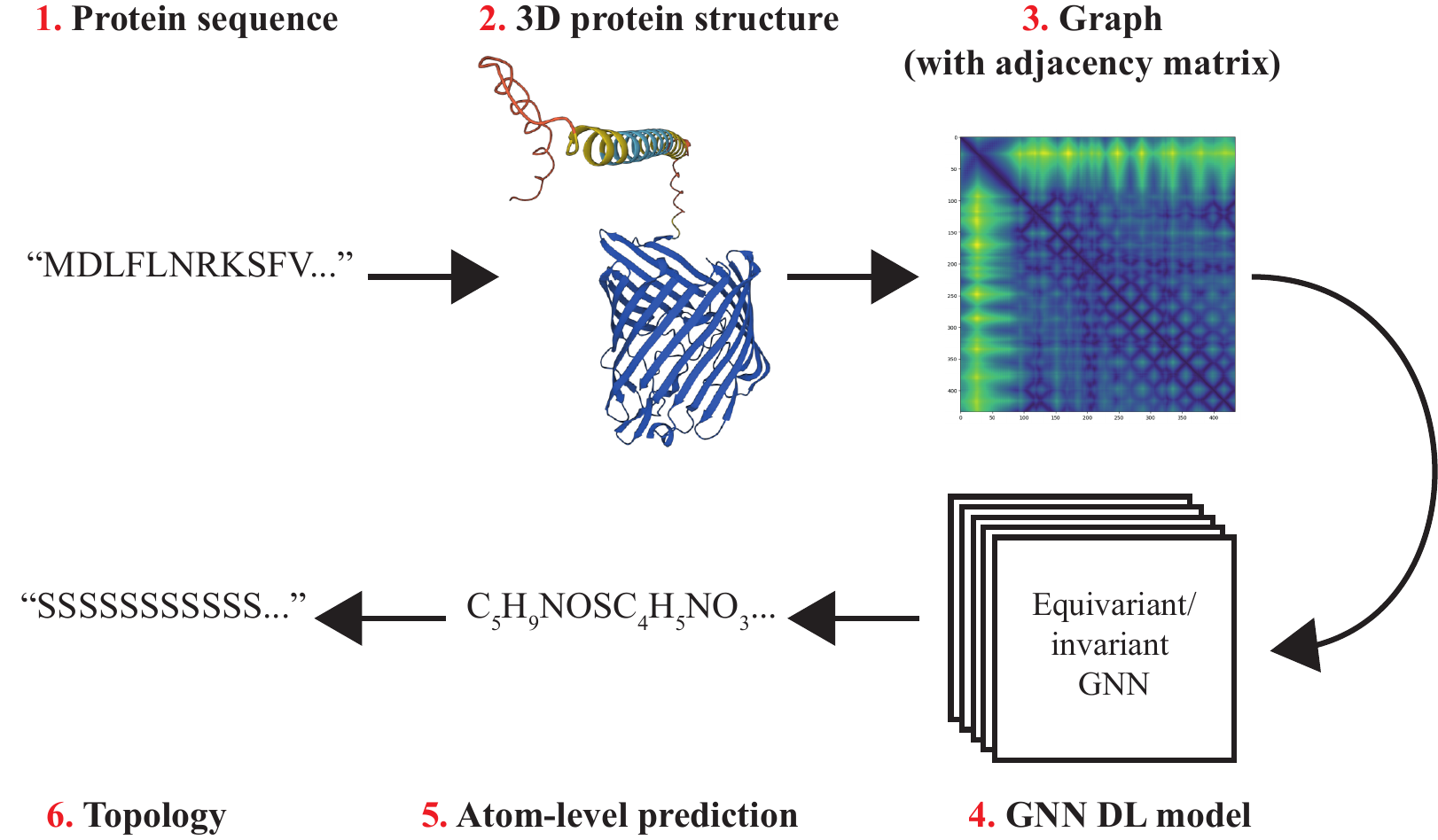}
    \caption{Methodology throughout the project. All blocks are numbered for later reference. The
3D protein structure in block 2 is from AlphaDB 3D viewer.}
    \label{fig:flow_diagram}
\end{figure}

Figure \ref{fig:flow_diagram} presents the general approach throughout this study. The key parts of the process are described in the subsequent sections.

\subsection{Dataset \& 3D protein structure}
\label{sec:data}

The 1D protein sequences are the same as in the DeepTMHMM study. The corresponding 3D structures are predicted by Alphafold and acquired from the Alphafold Data Base (AlphafoldDB) \cite{jumper2021highly, varadi2022alphafold}. Protein sequences without their 3D structures are excluded from the study, Table \ref{tab:data} provides an overview of which protein types are included and whether or not their 3D predictions are available from the AlphaDB.
\begin{table}[H]
\fontsize{10}{12}\selectfont
\centering
\caption{Number and name of each protein type used in the DeepTMHMM study. \textit{TM}: Transmembrane, \textit{SP}: Signal Peptide. The numbers without parentheses indicate sequences with both 1D sequences and 3D structures available while the numbers within parentheses indicate proteins with only 1D sequences.}
\vspace{2mm}
\label{tab:data}
\resizebox{\columnwidth}{!}{%
\begin{tabular}{cccccc}
\toprule
\hline
\textbf{\begin{tabular}[c]{@{}c@{}}Protein \\ type\end{tabular}}             & alpha TM  & alpha SP+TM & beta barrel & Globular      & SP+Globular \\ \hline
\textbf{Amount}                                                              & 383 (387) & 102 (106)   & 82 (82)     & 1,982 (2,000) & 994 (1,000) \\ \hline
\bottomrule
\end{tabular}
}
\end{table}
The 3D protein structures are stored in \texttt{.pdb} files. Each \texttt{.pdb} file contains the protein 1D sequence, the atoms of the protein, the atom coordinates, and the prediction uncertainties. A graph in the form of an adjacency matrix can then be constructed based on the atom coordinates (see block \textbf{3} in Figure \ref{fig:flow_diagram}). The atom types include carbon (C), nitrogen (N), oxygen (O), and sulfur (S) while hydrogen (H) is omitted probably due to its relatively low importance. Upon data inspection, we observed that every residue (amino acid in bound form) would start with the nitrogen atom in the graph. This was later exploited for the decoder implementation used between blocks 5 and 6 in Figure \ref{fig:flow_diagram}.

\subsection{Graph Neural Network}
\label{sec:GNN}

\begin{figure}[H]
    \centering
    \includegraphics[width=\linewidth]{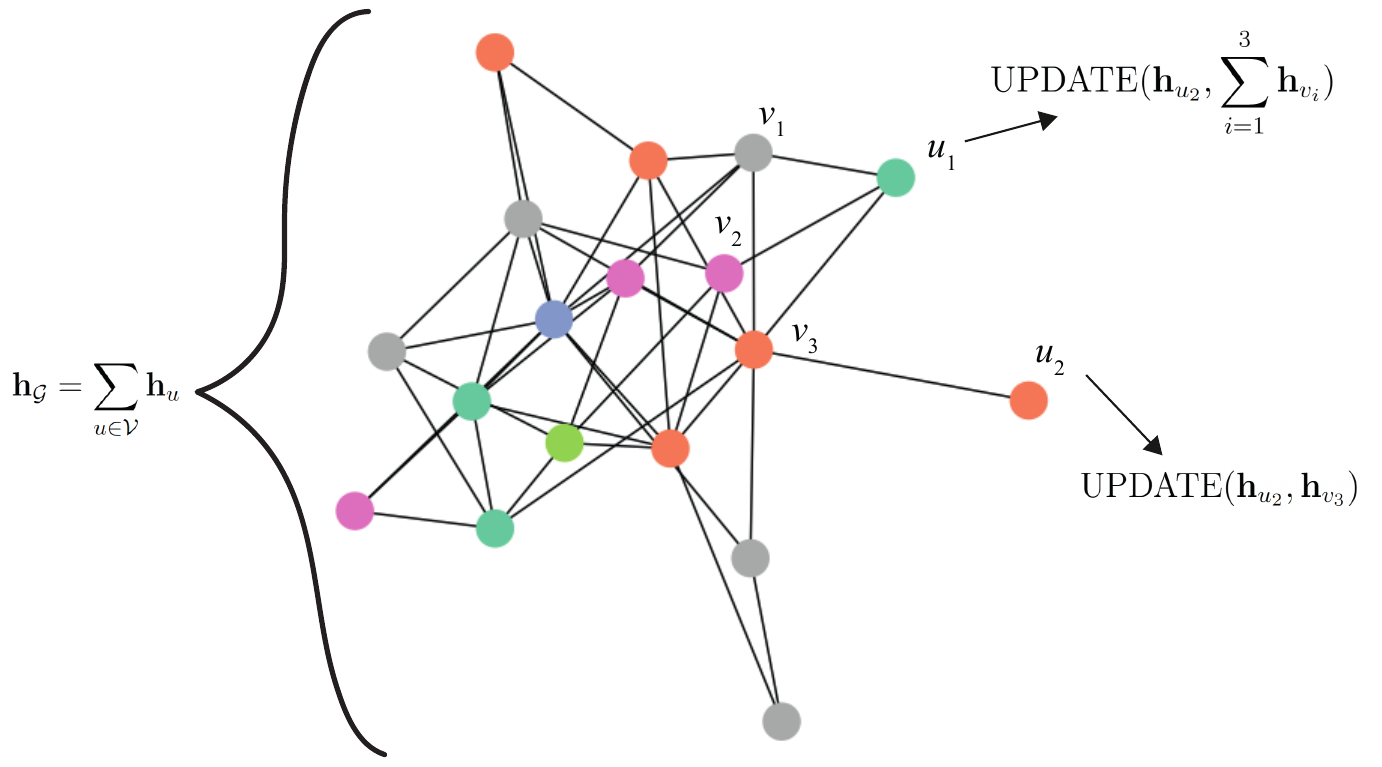}
    \caption{Random graph illustrating the basic message passing of GNN. Notice that the AGGREGATE operation is part of the UPDATE function (indicated by a sum) and depends on the number of neighbors. The graph embedding is then a sum of all the node embeddings.}
    \label{fig:graph}
\end{figure}

Generally, specific models are needed to process the graphs. A Graph Neural Network (GNN) is one of the models that takes a graph as input and returns as outputs a set of the so-called \textit{node embeddings} corresponding to all the nodes of the graph and a \textit{graph embedding} for the whole graph \cite{hamilton2020graph}. 
More generally, the GNN uses an AGGREGATE and an UPDATE function to perform the embedding calculations:
\begin{equation}
    m_{\mathcal{N}(u)} = \mathrm{AGGREGATE}^{(k)}(\{\bm{\mathrm{h}}_v^{(k)},	\forall v \in \mathcal{N}(u)\}),
\end{equation}
\begin{equation}
    \mathrm{UPDATE}(\bm{\mathrm{h}}_u, \bm{\mathrm{m}}_{\mathcal{N}(u)}) = \sigma\,(\bm{\mathrm{W}}_{\mathrm{self}}\bm{\mathrm{h}}_u + \bm{\mathrm{W}}_{\mathrm{neigh}}\bm{\mathrm{m}}_{\mathcal{N}(u)}).
\end{equation}
where $u$ is the current node being calculated, $\mathcal{N}(u)$ is the neighborhood of $u$. $v$ is a neighbor node to $u$ and thus belongs to $\mathcal{N}(u)$, $\bm{\mathrm{h}}_u$ is a so-called hidden node embedding, $\sigma$ is a nonlinear function that can be e.g. a ReLU function while $\bm{\mathrm{W}}_{\mathrm{self}}$ and $\bm{\mathrm{W}}_{\mathrm{neigh}}$ are trainable parameters \cite{hamilton2020graph}. In the context of protein topology, we focus on the node embeddings. Figure \ref{fig:graph} shows how the node embeddings and the graph embedding are obtained from the graph using the message passing mechanism.

The output of an \textit{invariant} GNN always has the same order while the output of an \textit{equivariant} GNN has the same order as the input order. In this study, we applied the state-of-the-art GNNs such as SchNet (invariant), EGNN (equivariant), and GCPNet (equivariant) to the topological task. However, SchNet was the only GNN that we managed to get meaningful results from.

\subsection{Atom-to-residue translation}
\label{sec:translation}
\begin{figure}[H]
    \centering
    \includegraphics[width=\linewidth]{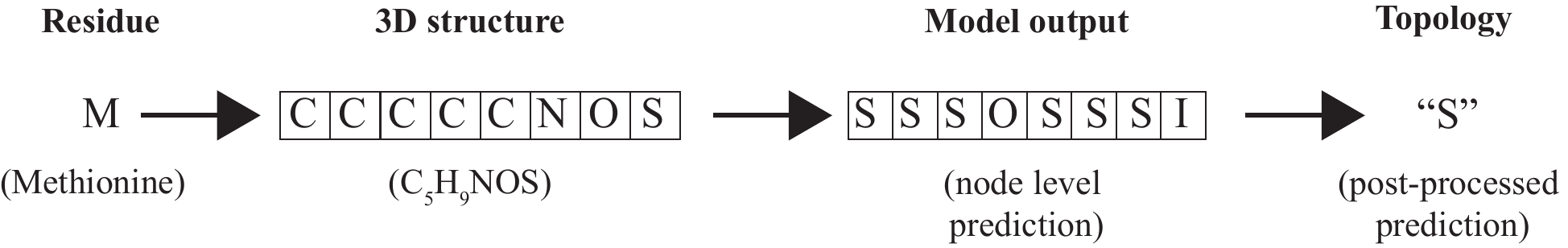}
    \caption{Major voting approach used in this study. The final topological prediction is based on the majority of the atom-level topological predictions.}
    \label{fig:prediction}
\end{figure}

GNN outputs node embeddings correspond to each atom of the protein graph. This means that the output dimension is much greater compared to the topological labels, which are on the residue-level. To translate the embeddings from the atom-level to the residue-level, we devised a \textit{major voting} approach, which is explained in Figure \ref{fig:prediction}. The gist of major voting is that all atoms give a topological prediction. The topology that is agreed upon by all the atoms within the residue would be selected to represent the topology of the residue. In case of a tie between the atom predictions, either candidate was selected to represent the residue-level topology.

In practice, the atoms within a residue were grouped using the protein sequence and the residue chemical formulas. The starting point of every residue was found by searching the nitrogen atom, which always served as the starting atom of each residue in the 3D graph as mentioned earlier. Subsequently, we could gather the other atoms within the residue to vote for the residue topology. More importantly, we created the atomic topology labels ourselves to align with the GNN predictions in the decoding stage before output.

Another approach is to train the GNN using the $\alpha$-carbon embedding from each residue \cite{anonymous2023evaluating}. An $\alpha$-carbon is the central atom linking to an amino group and a carboxyl group within an amino acid \cite{alberts2015essential}. In this way, the output dimension based on the $\alpha$-carbons would have the same dimension as the protein sequence. However, we only achieved similar or worse performance by using $\alpha$-carbons compared to using the aforementioned major voting approach.

\subsection{Training data distribution and Baseline model}
\label{sec:baseline}
The residue classes are \textit{signal peptide (S)}, \textit{inside cytosol (I)}, 
\textit{alpha membrane part (M)}, \textit{beta membrane part (B)}, \textit{periplasm (P)} and \textit{outside cell (O)} \cite{hallgren2022deeptmhmm}. This study uses a 5-fold cross-validation (CV) to assess the model performance. We used the same splits from the DeepTMHMM study in such a setup so that the model was trained on the first three sets, validated on the fourth set, and tested on the fifth set. This process had been repeated five times and each time the model was validated and tested on different sets. The hyper-parameter tuning was conducted only on the first fold setup because no major performance difference was observed with the other folds.

\begin{figure}[H]
    \centering
    \includegraphics[width=0.8\linewidth]{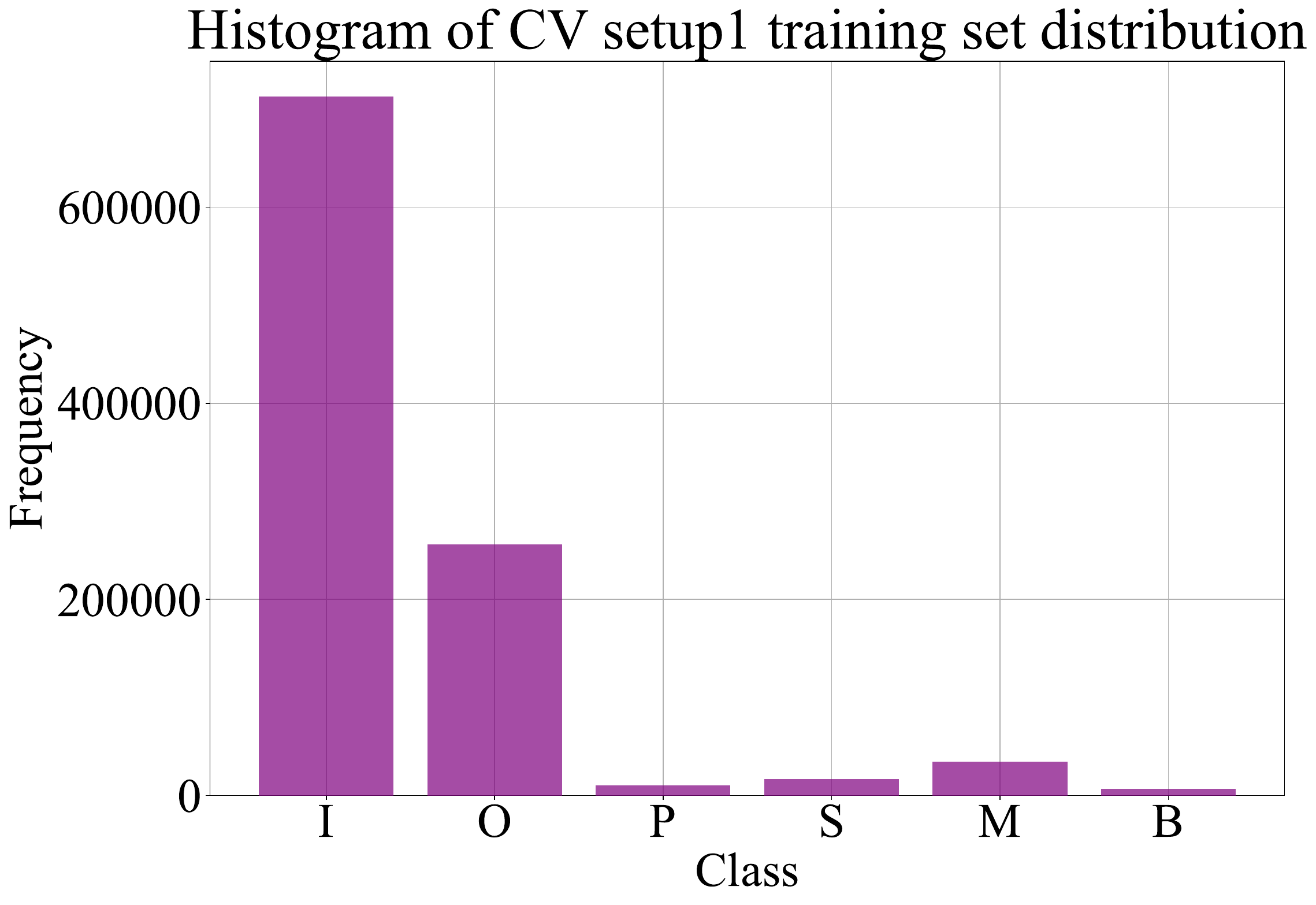}
    \caption{Class distribution in the first fold of the 5-fold cross-validation. This distribution indicates that most protein parts intracellular proteins. The distributions in the other CV folds are very similar (not shown).}
    \label{fig:data_distri}
\end{figure}

The model performance was compared to a baseline classifier, which predicted all test observations as belonging to the most frequently appearing class in the training set. We observed that in all the CV folds the class \textit{inside cytosol (I)} was the most frequently appearing class and the class distribution was very similar for all the CV folds. The comparison was conducted using McNemars test \cite{Intro_ML}. Figure \ref{fig:data_distri} shows the distribution of all the label classes in the first fold setup. 

\subsection{Model specification}
\label{sec:model_spec}
We applied a SchNet model with 6 layers, a hidden embedding size of 128, 128 convolutional filters, a maximum number of 32 neighbor nodes to be included for the message passing, and a dropout set to 0.8 right before the fully connected layers. A batch size of 1 was chosen because this gave the best and most stable performance. The Adam optimizer was used with a learning rate of $3\cdot 10^{-4}$ and $L2$ regularization of $1\cdot 10^{-4}$. Additionally, an exponential \textit{learning rate scheduler} was applied with an exponential decay rate of 0.1. The learning scheduler was applied when the validation loss stopped decreasing, which was detected using \textit{early stopping}.

In \cite{bernhofer2022tmbed}, M. Bernhofer and B. Rost applied a Gaussian smoothing filter with a kernel size of 7 and standard deviation ($\sigma$) of 1 to the node embeddings for each class before converting them into class probabilities. The intention of Gaussian smoothing during training was to eliminate the irregular and sudden appearance of residues. As for our study, we also observed that the residue topology predictions would sometimes change abruptly. Hence, we applied a Gaussian filter during the training and validation stage with a kernel size of 29 and $\sigma$ of 5. Please notice that our input graph nodes are at atom-level, which means that a larger kernel size and $\sigma$ are needed to achieve the same smoothing effect on residue-level.

We had experimented with all the existing pre-trained weights provided by the \textit{ProteinWorkshop} framework \cite{protein_workshop}, which were from the tasks \textit{inverse folding}, \textit{predicted
local distance difference test (pLDDT) prediction}, \textit{sequene denoising}, \textit{structure denoiseing} and \textit{torsional denoising}. Unfortunately, none of these weights improved the training.  
The Python library \texttt{wandb} was employed to monitor the learning and loss curves. The models were trained using Tesla A100 (32 GB and 80 GB) owned by the Technical University of Denmark (DTU) Compute department.

\section{Results}
\label{sec:results}

\begin{figure}[H]
    \hspace{-1cm}
    \begin{subfigure}[b]{1.1\linewidth}
    \includegraphics[width=\linewidth]{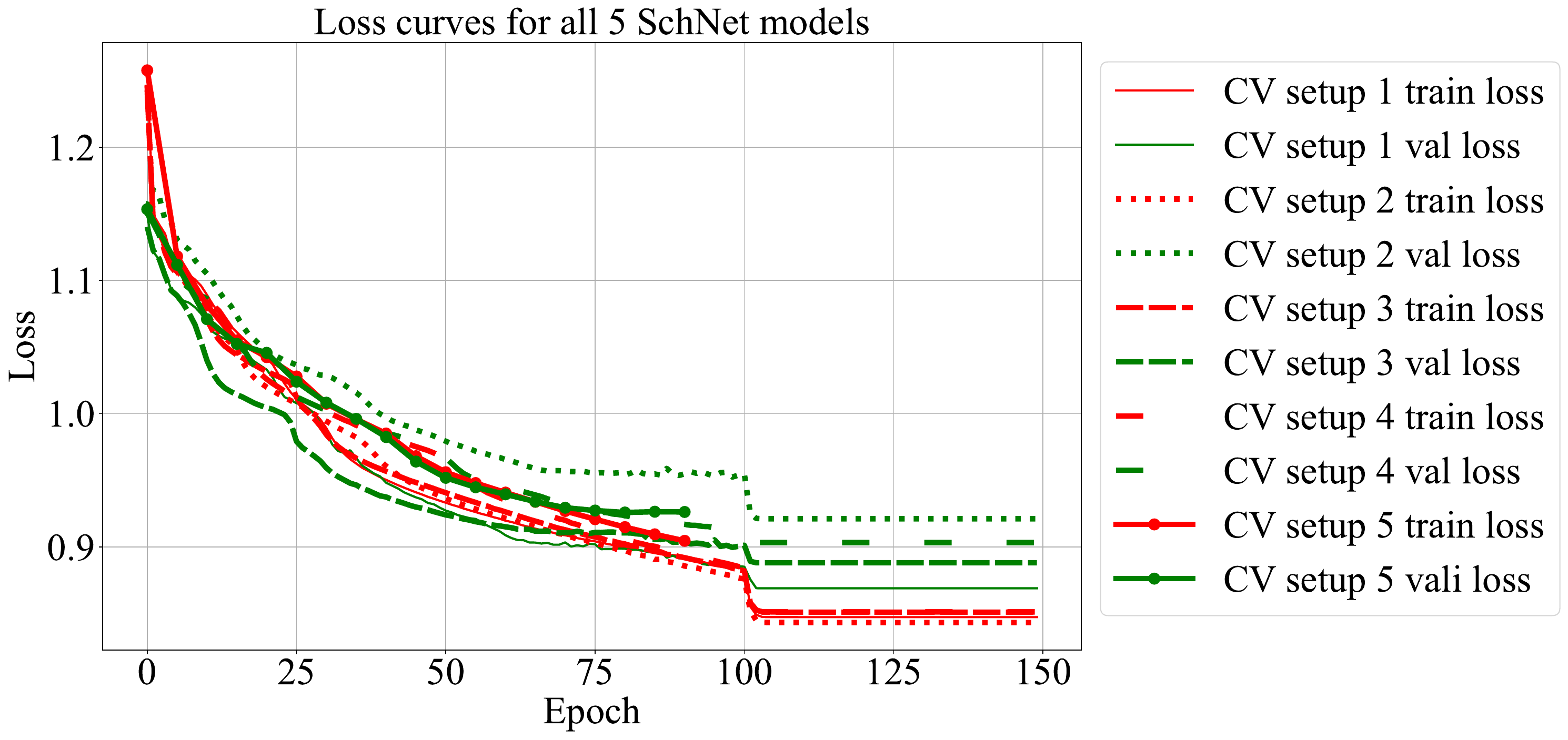}
    \end{subfigure}
    \begin{subfigure}[b]{1.1\linewidth}
    
    \hspace{-1cm}
    \includegraphics[width=\linewidth]{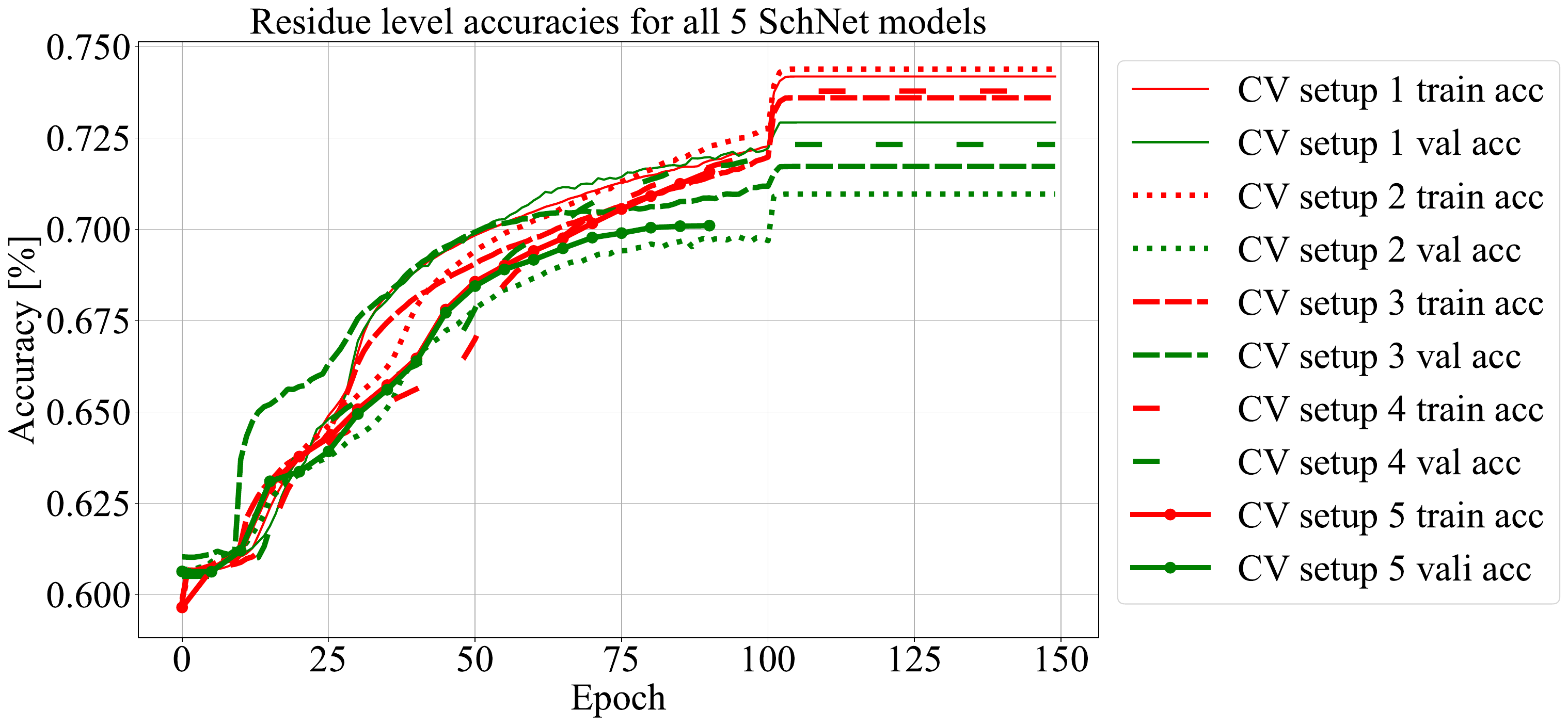}
    \end{subfigure}
    
    \caption{a): Loss curves for the 5 SchNet models trained and validated on each of the five CV folds. b): Accuracy curves for the 5 SchNet models trained and validated on each of the five CV folds. The abrupt jump around epoch 100 is caused by the exponential decaying learning scheduler. Please notice that the training for CV setup 5 was early-stopped around epoch 90.}
    \label{fig:loss_and_acc}
\end{figure}

This section presents the main results obtained from SchNet and other relevant test results.
Figure \ref{fig:loss_and_acc} shows the training and validation losses and residue-level accuracies for the five SchNet models. The residue-level accuracies are evaluated by directly comparing them to the class labels. It can be seen that all the loss and accuracy curves have similar tendencies, indicating that the variability is small in each CV setup. 

\subsection{Topological evaluation}

The evaluation criteria for two same protein topologies are defined according to \cite{hallgren2022deeptmhmm} as the following:
\begin{itemize}
    \item The N-terminal topology must be the same.
    \item The predicted labels and the ground truth labels must overlap with 5 residues for $\alpha$-helices and 2 for $\beta$-strands (sub-part of $\beta$-barrel).
\end{itemize}

The test results are gathered from the five test sets of the 5-fold CV, so the topological predictions are evaluated using the above-mentioned criteria for all the five protein types. 
In Table \ref{tab:topo_result}, it can be seen that only very few topologies have been predicted for \textit{Globular} and \textit{SP+Globular} by the trained SchNet models. However, the average correctly predicted residues per protein is still much higher for SchNet than for the baseline. The average correctly predicted residues per protein are also calculated for the DeepTMHMM predictions to show how far away our models are from achieving a similar performance as DeepTMHMM.

\begin{table}[H]
\centering
\caption{Topological and residue-level evaluation performed on the protein types \textit{alpha TM}, \textit{SP+TM}, \textit{beta barrel}, \textit{Globular} and \textit{SP+Globular} presented in Table \ref{tab:topo_result}. Please notice that the overlap criterion is set to be 5 by default for the topological evaluation of \textit{Globular} and \textit{SP+Globular} as these proteins contain both $\alpha$-helices and $\beta$-sheets \cite{alberts2015essential}.}
\resizebox{\columnwidth}{!}{%
\begin{tabular}{c|cccc}
\toprule
\hline
\textbf{\diagbox[width=10em]{Metrics}{Protein types}}                                                              & alpha TM & alpha SP+TM & beta barrel & Globular \& SP+Globular \\ \hline
\textbf{\begin{tabular}[c]{@{}c@{}}Correct topology \\  by SchNet [\%]\end{tabular}}                                                   & 0      & 0         & 0  & 3.6        \\ \hline
\textbf{\begin{tabular}[c]{@{}c@{}}Correct topology \\ by baseline [\%]\end{tabular}}                                                    & 0      & 0         & 0  & 0       \\ \hline
\textbf{\begin{tabular}[c]{@{}c@{}}Correct topology \\ by DeepTMHMM [\%]\end{tabular}}                                                    & 83.5      & 92.5         & 80.3  & *95.6       \\ \hline
\textbf{\begin{tabular}[c]{@{}c@{}}Average correct residue \\ labels per protein\\ by SchNet [\%]\end{tabular}} & 59.5     & 50.8        & 26.4     & 75.2   \\ \hline
\textbf{\begin{tabular}[c]{@{}c@{}}Average correct residue \\ labels per protein \\ by baseline [\%]\end{tabular}} & 39.8     & 17.3        & 0.0   & 66.6     \\ \hline
\textbf{\begin{tabular}[c]{@{}c@{}}Average correct residue \\ labels per protein \\ by DeepTMHMM [\%]\end{tabular}} & 88.4     & 94.5        & 87.6   & 96.9    \\ \hline
\bottomrule
\multicolumn{5}{l}{\footnotesize{*Calculated by our own using an overlap of 5 residues.}}\\
\end{tabular}
}
\label{tab:topo_result}
\end{table}

\subsection{Comparison between baseline and SchNet using McNemars test}

To verify the validity of our trained GNNs, we compared the baseline with the SchNet models using the McNemars test. In Table \ref{tab:McNemars}, the comparison is presented for each of the 5-fold CV setups. The SchNet models have better residue-level accuracies compared to the baseline. Meanwhile, 0 is always outside the 95 \% confidence intervals (CIs), and the p-values are always strongly significant as well. Overall, all this indicates that the SchNet models are better suited for topological prediction than the baseline. 

\begin{table}[H]
\caption{McNemars test and the corresponding 95 \% confidence intervals (CIs) between the baseline and the trained SchNet model for each CV setup. The trained SchNet is always superior regardless of which setup.}
\resizebox{\columnwidth}{!}{%
\begin{tabular}{c|ccccc}
\toprule
\hline                                                                                             \textbf{\diagbox[width=10em]{Metrics}{CV fold}}   & \textbf{CV setup 1}  & \textbf{CV setup 2}  & \textbf{CV setup 3}  & \textbf{CV setup 4}  & \textbf{CV setup 5}  \\ \hline
\textbf{\begin{tabular}[c]{@{}c@{}}Baseline residue-level \\ accuracy {[}\%{]}\end{tabular}}                 & 68.0                 & 69.3                 & 69.4                 & 67.5                 & 69.3                 \\ \hline
\textbf{\begin{tabular}[c]{@{}c@{}}SchNet residue-level \\ accuracy {[}\%{]}\end{tabular}}                   & 75.2                 & 76.9                 & 75.5                 & 76.3                 & 75.6                 \\ \hline
\textbf{\begin{tabular}[c]{@{}c@{}}Comparison between \\ baseline and SchNet \\ (p-value)\end{tabular}}   & 0.00                 & 0.00                 & 0.00                 & 0.00                 & 0.00                 \\ \hline
\textbf{\begin{tabular}[c]{@{}c@{}}Comparison between \\ baseline and SchNet \\ (95 \% CI )\end{tabular}} & {[}-0.074, -0.071{]} & {[}-0.078, -0.075{]} & {[}-0.062, -0.059{]} & {[}-0.089, -0.086{]} & {[}-0.063, -0.061{]} \\     
\hline 
\bottomrule
\end{tabular}
}
\label{tab:McNemars}
\end{table}

Finally, it is noteworthy that we initially used a hold-out approach to train and validate the GNN model on only \textit{Globular} and \textit{SP+Globular} proteins and tested on the remaining three protein types. However, the result was worse compared to the 5-fold CV.

\subsection{Results from $\alpha$-carbon method and other models}

\begin{figure}[htbp]
    \includegraphics[width=\linewidth]{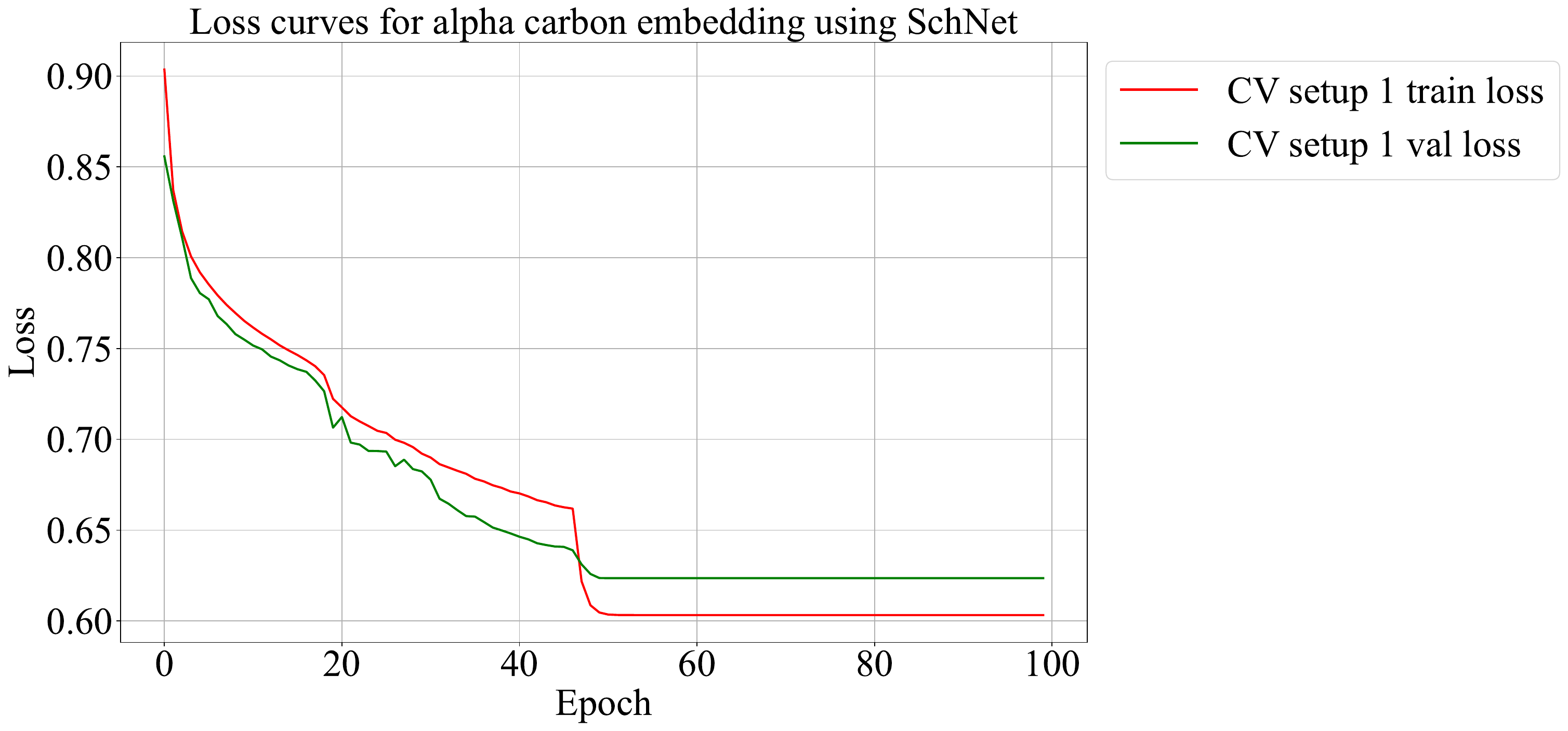}
    \includegraphics[width=\linewidth]{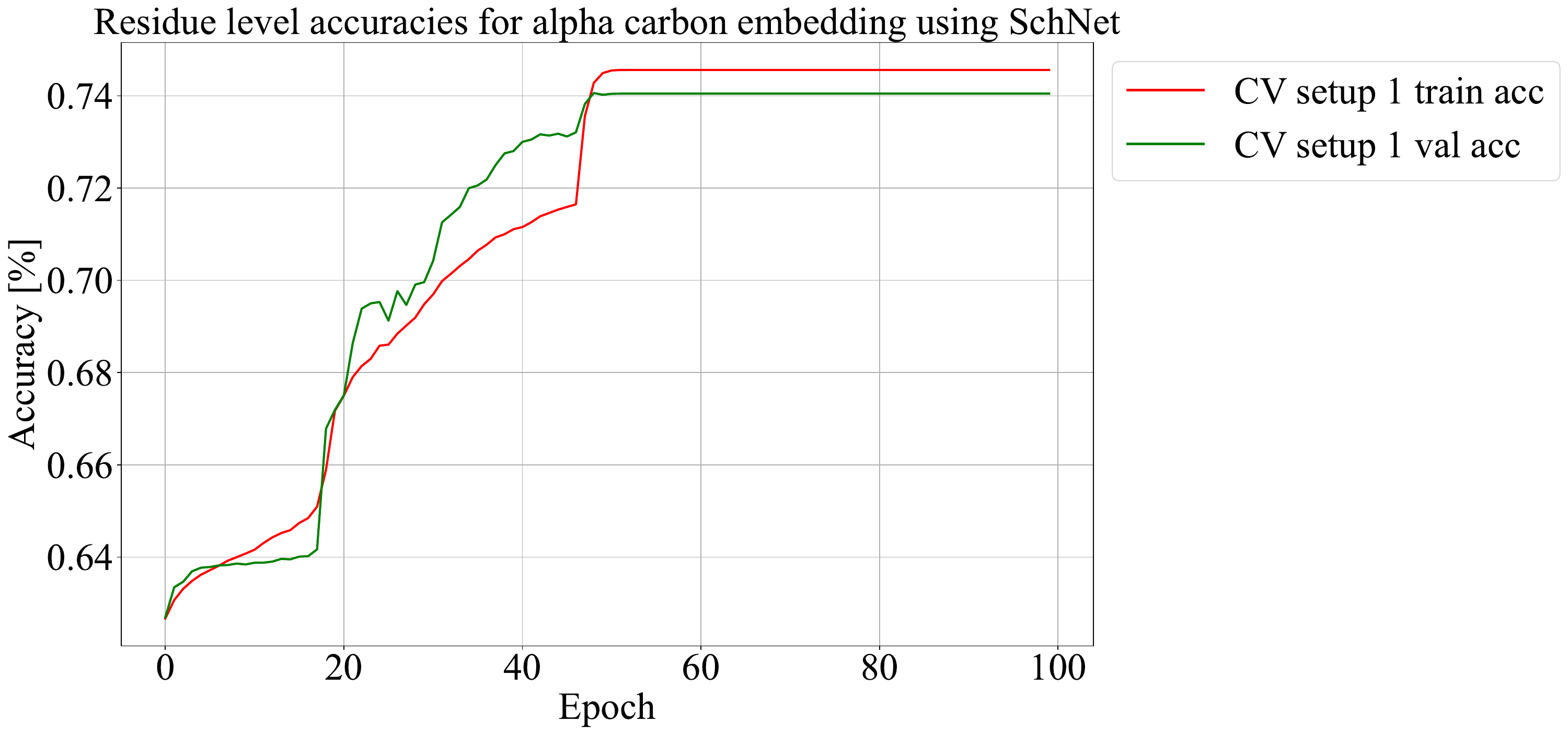}

    \caption{SchNet trained and validated on CV setup 1 using $\alpha$-carbon approach with Gaussian smoothing. The hyper-parameters are fine-tuned and an exponential learning schedule is applied.}
    \label{fig:loss_and_acc_CA_carbon}
\end{figure}

\begin{figure}[htbp]
\centering
    \includegraphics[width=\linewidth]{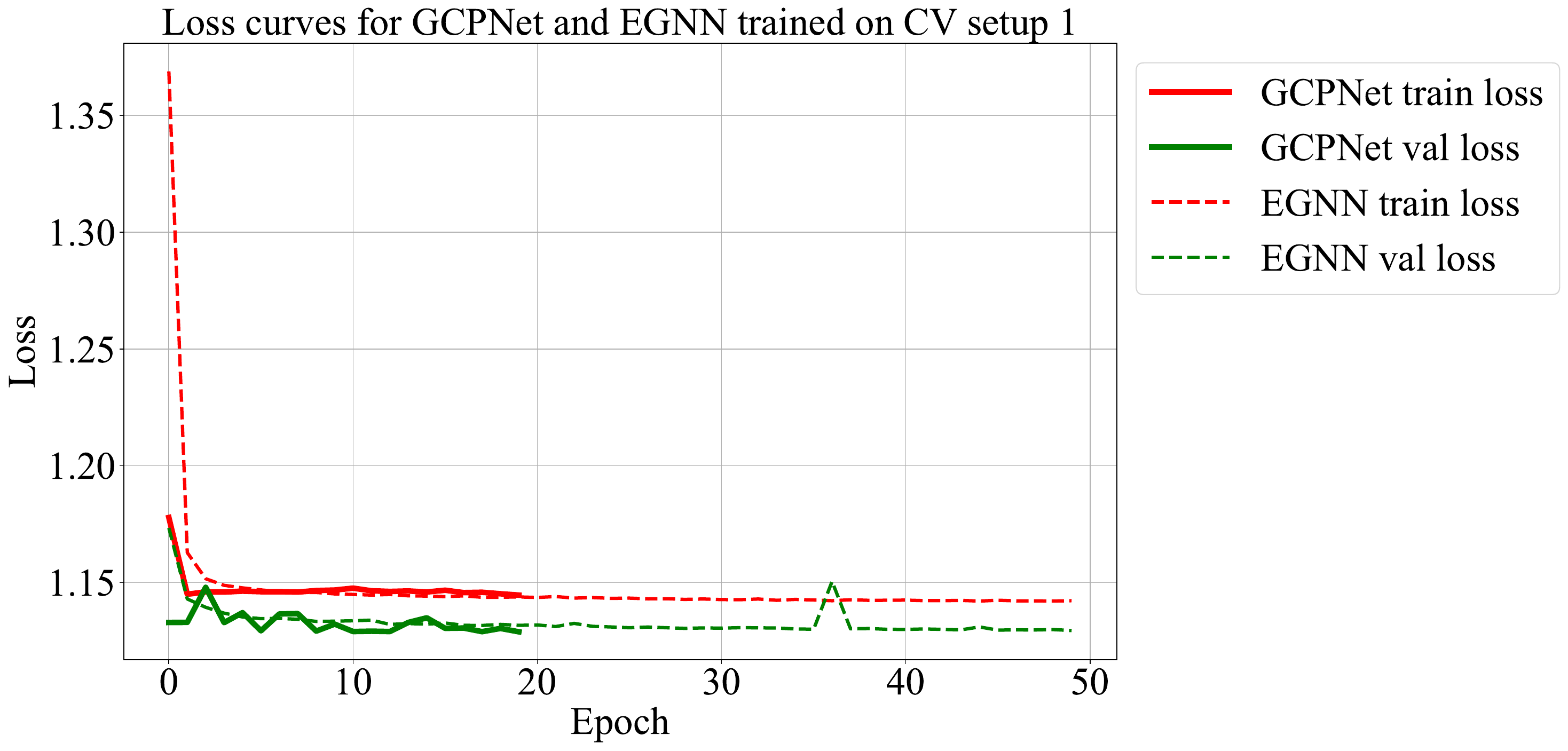}
    \includegraphics[width=\linewidth]{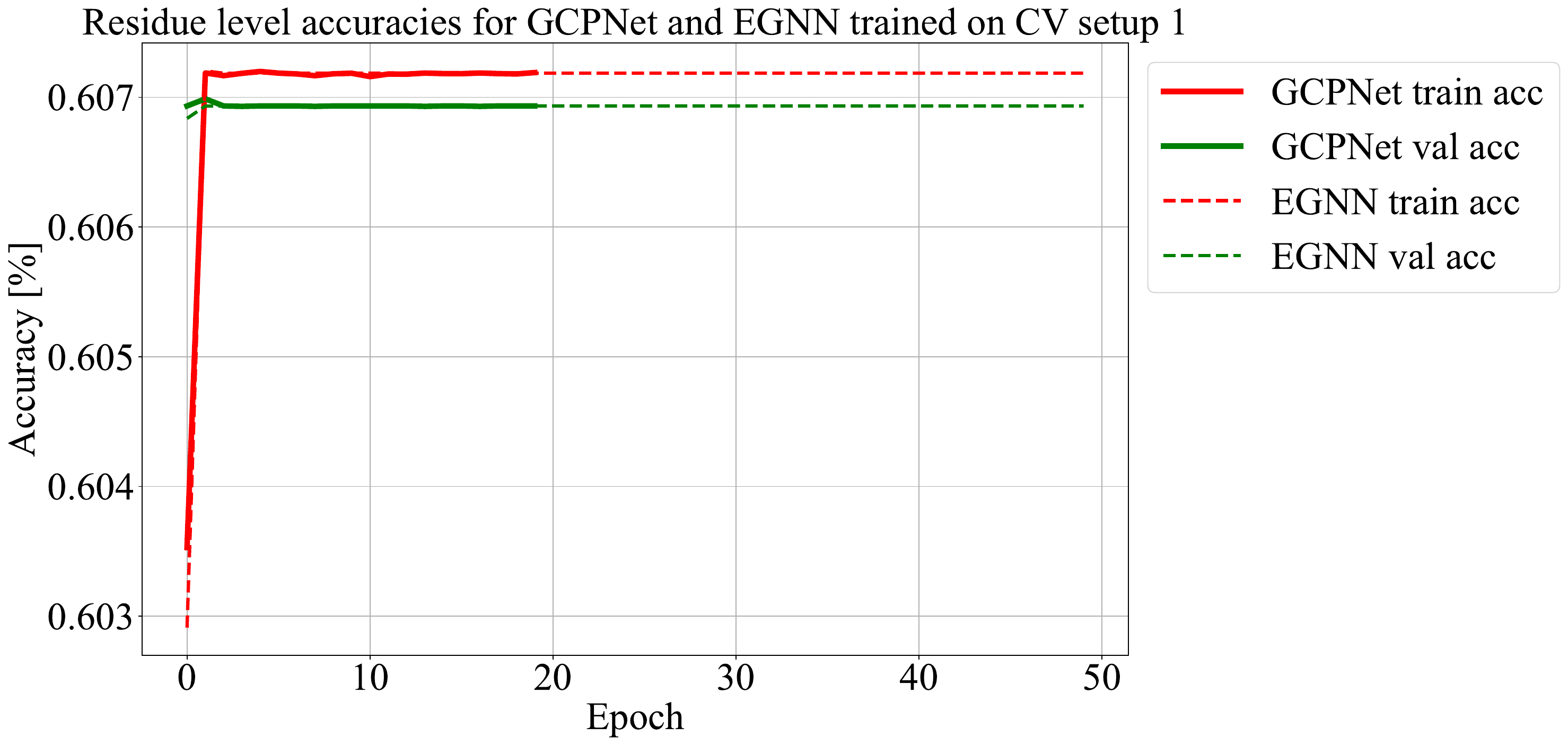}
    
    \caption{Training and validation curves of GCPNet and EGNN trained and validated on CV setup 1 with the proposed major voting method. Both models seem to have failed to learn the representation from the CV setup 1 dataset. Gaussian smoothing is applied to the models.}
    \label{fig:loss_and_acc_other_models}
\end{figure}

Figure \ref{fig:loss_and_acc_CA_carbon} shows the SchNet performance trained on CV setup 1 using the previously mentioned $\alpha$-carbon approach. While the loss curves converge at a lower value compared to the major voting approach, the actual number of the correctly predicted residue and the number of correct topological predictions do not improve. These results are not included to avoid confusion. Figure \ref{fig:loss_and_acc_other_models} shows the performance of GCPNet and EGNN trained on CV setup 1. It can be seen that both models fail to learn the topology from the data because they have approximately constant losses and accuracies. It has to be emphasized that we were unable to train GCPNet and EGNN with large batch sizes due to their demanding memory requirements. The numerical prediction results from the two models are below the baseline predictions and thus not presented here.

\section{Discussion}
\label{sec:discussion}

The comparison between the trained SchNet models and the baseline model shows that SchNet models are better at determining the residue-wise topology classes. Nonetheless, the overall topological prediction is far from satisfactory. One of the main reasons for this could be that the model was trained from scratch without using any pre-trained weight. The current state-of-the-art models like DeepTMHMM and TMbed are based on sophisticated protein language models (pLMs), which are pre-trained on billions of protein sequences \cite{bernhofer2022tmbed}. Furthermore, the time and computational constraints prevented us from experimenting with other GNNs, which might be more well-suited for the topology task. Finally, the amounts of the protein types \textit{alpha TM}, \textit{alpha SP+TM} and \textit{beta barrel} are way less compared to \textit{Globular} and \textit{SP}+\textit{Globular} as shown in Table \ref{tab:data}. This data imbalance most likely has contributed to the fact that the model is better at predicting the topologies for \textit{Globular} and \textit{SP}+\textit{Globular} than for \textit{alpha TM}, \textit{alpha SP+TM} and \textit{beta barrel}. Finally, it is important to emphasize that the 3D structures predicted by Alphafold also contain uncertainties, which inevitably would limit the performance of the models that are trained upon AlphaDB.

\section{Conclusion}
\label{sec:conclusion}
In this study, we combined knowledge from biology, dynamic programming, and deep learning to apply state-of-the-art GNNs such as SchNet, EGNN, and GCPNet to identify protein topology. Specifically, we developed the major voting method to infer the residue-wise topological labels by using a consensus of the atom-level topological predictions. We also explored another well-known method based on extracting the feature from the $\alpha$-carbons, but this method did not show to improve the performance compared to our major voting approach. Within the computational budget, only SchNet gave the best results although its performance was still far away from the state-of-the-art protein models such as DeepTMHMM and TMbed. Nonetheless, with more pre-trained GNNs becoming available concerning the topological task, the performance would certainly be improved soon.

\section{Acknowledgment}
\label{sec:acknewledgment}
This project was accomplished in fall 2023 during the course 02456 Deep Learning at DTU. We thank DTU Compute for providing the computational resources. In addition, we thank Felix Teufel for his supervision, advice, and provision of relevant materials during the project.

\balance
\bibliographystyle{IEEEbib}
\bibliography{refs}

\end{document}